\documentclass[10pt,twocolumn,letterpaper]{article}

\usepackage{wacv}
\usepackage{times}
\usepackage{epsfig}
\usepackage{graphicx}
\usepackage{amsmath}
\usepackage{amssymb}

\usepackage{hyperref}
\usepackage{url}
\usepackage{verbatim}
\usepackage{caption}
\usepackage{subcaption}
\usepackage{soul}
\usepackage{algorithm}
\usepackage{algorithmic}

\usepackage{float}
\floatstyle{plaintop}
\wacvfinalcopy % *** Uncomment this line for the final submission

\def\wacvPaperID{58} % *** Enter the wacv Paper ID here

\ifwacvfinal\fi
\begin{document}

%%% Title Page for unofficial publishing %%%

\begin{titlepage}
\centering{
  \vspace*{\fill}
  \begin{center}
    {\Huge Linear-time Online Action Detection From 3D Skeletal Data Using Bags of Gesturelets}\\[0.4cm]
    {\large Moustafa Meshry, Mohamed E. Hussein and Marwan Torki}\\[1cm]
    {\LARGE This paper is accepted for publication at the IEEE Winter Conference on Applications of Computer Vision, Lake Placid, NY, 2016\\[0.3cm]}
    {\large Copyright @ IEEE 2016\\}
  \end{center}
  \vspace*{\fill}
}
\end{titlepage}

%%%%%%%%% TITLE
\title{Linear-time Online Action Detection From 3D Skeletal Data Using Bags of Gesturelets}

% Authors at the same institution
\author{Moustafa Meshry$^1$ \hspace{2cm} Mohamed E. Hussein$^{1,2}$\thanks{Mohamed E. Hussein is currently an Assistant Professor at Egypt-Japan Univeristy of Science and Technology, on leave from his position at Alexandria University.} \hspace{2cm} Marwan Torki$^1$ \\
$^1$ Dept. of Computer and Systems Engineering, Alexandria University, Egypt\\
$^2$ Dept. of Computer Science and Engineering, Egypt-Japan University of Science and Technology\\
{\tt\small moustafa.meshry@alexu.edu.eg, mohamed.e.hussein@ejust.edu.eg, mtorki@alexu.edu.eg}
}
% Authors at different institutions
% \author{First Author \\
% \and
% Second Author \\

\maketitle
\ifwacvfinal\thispagestyle{empty}\fi

%%%%%%%%% ABSTRACT
\begin{abstract}
  Sliding window is one direct way to extend a successful recognition system to 
  handle the more challenging detection problem. 
  While action recognition decides only whether or not an action is present in a 
  pre-segmented video sequence, action detection 
  identifies the time interval where the action occurred in an unsegmented video 
  stream. 
  Sliding window approaches can however be slow as they maximize a classifier 
  score over all possible sub-intervals. 
  Even though new schemes utilize dynamic programming to speed up the search for 
  the optimal sub-interval, they require 
  offline processing on the whole video sequence. 
  In this paper, we propose a novel approach for online action detection based on 
  3D skeleton sequences extracted 
  from depth data. It identifies the sub-interval with the maximum classifier 
  score in linear time. 
  Furthermore, it is suitable for real-time applications with low latency.
\end{abstract}
\section{Introduction}
\label{sec:intro}
Human action detection at real-time has become a topic of increasing interest 
due to its wide practical use. Applications like Human-machine interaction, 
surveillance and gaming, all require accurate and low-latency action detection. 
Action detection on raw videos is difficult because it is first needed to 
localize a person in a scene full of objects and clutter, then try to recognize 
the type of action being performed. 
On the other hand, the recent low-cost depth sensors, like Microsoft Kinect, 
provided a more convenient way for data capture. The 3D positions of body joints 
can be estimated from depth maps at low-latency and with acceptable accuracy. 
Filtering out background clutter, it is now more adequate to perform action 
detection based on skeleton data. Recently, skeleton-based approaches to action 
recognition and detection have been widely adopted. 
While action recognition focuses on identifying the action label of 
pre-segmented video sequences, action detection tackles the more challenging 
problem of temporally localizing the action in an unsegmented stream of frames.

The main contribution of this paper is a novel approach for action detection from skeleton data, 
that we refer to as Efficient Linear Search (ELS). 
We show that a combination of simple components and specializing them towards skeleton-based action
detection can achieve state-of-the-art results and overcome the limitations of similar approaches.
The proposed approach is flexible: it can be used with a wide class of 
classifier functions 
and with different types of action local descriptors. 
As a byproduct contribution, we propose a simple skeleton-based local descriptor 
that, when used in a 
simple bag-of-features model, produces state-of-the-art results on different 
datasets.
% \footnote{Before the camera ready version, we found~\cite{wu2014leveraging} who report better detection results than ours on the MSRC-12 dataset.}
The proposed framework works online and is suitable for real-time applications. Moreover, it can 
be used for 
real-time video segmentation, since it specifies both the start and end frames 
of the action. 

The rest of this paper is organized as follows: section~\ref{sec:relWork} gives 
an overview about recent related work in the literature. We show the used action representation
and our proposed descriptor in section~\ref{sec:mainIdea}. We, then, explain our Efficient Linear
Search approach in section~\ref{sec:els}. Experimental evaluation is presented in~\ref{sec:experiments}.
And finally, we conclude in~\ref{sec:conclusion}.
% 
% =============================================================================
% 
\section{Related work}
\label{sec:relWork}
Forming suitable skeletal-based descriptors for action recognition has been the focus of many recent research 
works~\cite{gowayyed2013histogram,anglesDesc,smij,sharaf2014real,wang2012mining,yu2015discriminative,zanfir2013moving,sss}. 
The objective is to facilitate the recognition task via a discriminative descriptor. 
Some of these descriptors capture both the pose of the skeleton and the kinematics at the same time on the frame level. 
For example, Nowozin \emph{et al.}~\cite{anglesDesc} proposed a local descriptor that 
uses 35 angles between triplets of joints, $\Theta (t_0)$, along with angular velocity, $\delta \Theta (t_0)$, to 
encode joints' kinematics. Joints angles are a powerful cue to skeleton pose. 
Moreover, they are invariant to body translation and rotation. 
Later, Zanfir \emph{et al.}~\cite{zanfir2013moving} proposed the Moving Pose descriptor, which 
captures both the body pose at one frame, as well as the speed and acceleration of body joints within 
a short time window centered around the current frame.

Another class of descriptors is focused on computing a fixed length descriptor for the whole action 
sequence, like ~\cite{gowayyed2013histogram,hussein2013human,lieGroup,wang2012mining}. 
Gowayyed \emph{et al.}~\cite{gowayyed2013histogram} used a 2D trajectory descriptor, called 
``Histogram of Oriented Displacements'' (HOD), where each displacement in the trajectory casts a 
vote, weighted by its length, in a histogram of orientation angles. 
Vemulapalli \emph{et al.}~\cite{lieGroup} modeled human actions as curves in a Lie group, since 
3D rigid body motions are members of the special Euclidean group. 
Wang \emph{et al.}~\cite{wang2012mining} used the 3D joints positions to construct a descriptor of 
relative positions between joints. 
However, descriptors on the whole sequence suffer from much higher dimensionality over 
those that were designed for the frame level. 
This higher dimensionality led sometimes to the need for feature selection as done in~\cite{smij,sharaf2014real}.

%Most of the focus on skeletal data was on action recognition. However, fewer works are focused on the online problem. 
While most of the focus on skeletal data was about action recognition, fewer works focused on 
the online problem. The trade-off between latency and accuracy was addressed in recent 
works~\cite{ellis2013exploring,anglesDesc,sharaf2014real,yu2015discriminative,zanfir2013moving,sss}. 
In ~\cite{anglesDesc} the notion of action points was first introduced and the detection problem was 
cast as a classification problem for every overlapping 35-frames intervals. The same notion of action 
points was utilized in the work of ~\cite{sharaf2014real}, but they could handle different scales. 
Other works, such as~\cite{yu2015discriminative,zanfir2013moving,sss} used the standard sliding 
window protocol for online action detection. Zhao \emph{et al.}~\cite{sss} proposed a feature 
extraction method, called ``Structured Streaming Skeleton'' (SSS), which constructs 
a feature vector for each frame using a dynamic matching approach. The SSS feature vectors are then used 
for detecting the start and end of actions. Zanfir \emph{et al.}~\cite{zanfir2013moving}used a modified 
kNN classifier to detect the start and end of actions. However, both ~\cite{zanfir2013moving,sss} cannot 
handle multi-scale actions, where the same action can be performed at different speeds. 
% 
% =============================================================================
% 
\section{Bag-of-gesturelets for action classification}
\label{sec:mainIdea}
The concept of local features first appeared in object detection in 
images~\cite{schmid1997local}. The main idea is that each object has a set of 
discriminative local features that, if appeared together, signify the existence 
of the object. Same concept applies to actions. For a specific action,
we can identify a set of key frames that best capture the discriminative poses 
of this action. 
However, skeleton poses alone cannot distinguish between some actions, e.g. 
standing up vs.~sitting down, since other information like the direction and 
speed of motion play an important role in identifying the action. So, 
differential quantities that describe joints' kinematics must be included in an 
action's local features. We then define a \textit{gesturelet} to be any general 
local feature that, for any frame, captures both the skeleton pose and 
kinematic information of body joints at this point in time. In the following, we first introduce our 
action representation as a bag of gesturelets. Then, we explain our local features descriptor for representing gesturelets.
%
%--------------------------------------------------------------------
%
\subsection{A bag-of-gesturelets representation}
\label{sec:representation}
Based on extracted gesturelets from action sequences, we make use of a 
bag-of-gesturelets (BoG) representation of human actions. We first extract features at each frame. 
Resulting descriptors are then clustered to 
produce a $K$-entry codebook. We then represent 
any action sequence or sub-sequence by its cluster histogram, where the 
histogram counts how many local features from each cluster index have occurred. In 
order to relax the assignment of each gesturelet to its representative cluster in 
the codebook, we apply \mbox{\emph{soft binning}}; letting each gesturelet cast a vote 
to its $m$ nearest clusters. The 
vote will be weighted according to the distance between the gesturelet and the 
corresponding cluster, so that closer clusters get higher weight. Implementation details for soft binning
are presented in section~\ref{sec:implementationdetails}.

The BoG representation is necessary for our approach for action detection, as we show in section~\ref{sec:els}. 
However, as shown in section~\ref{sec:exp_recognition}, it is also very effective for action recognition. 
This BoG representation is independent of the choice of the local descriptor used to represent a gesturelet. 
Possible descriptors that capture both pose information and joints kinematics 
are~\cite{anglesDesc,sharaf2014real,yu2015discriminative,zanfir2013moving}.
% 
%--------------------------------------------------------------------
%
\begin{figure*}[t]
  \centering
      \includegraphics[scale=0.55, trim={0 0.5cm 0 0},clip]{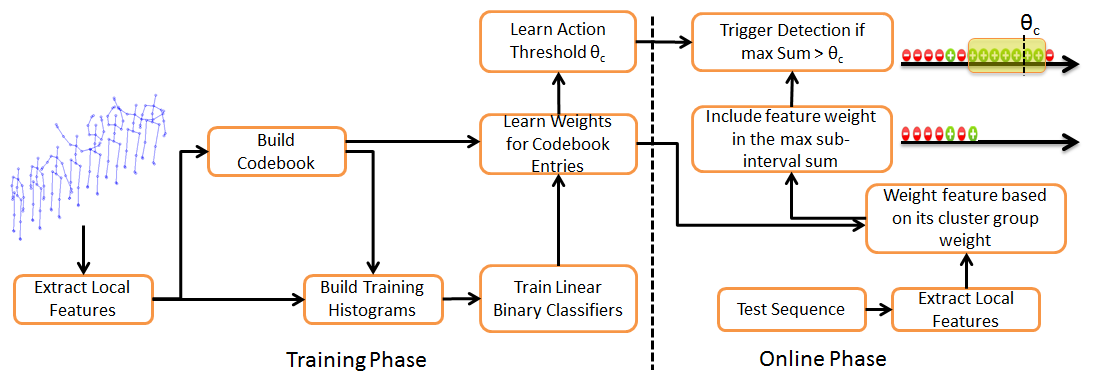}
  \caption{\small Efficient Linear Search Overview. The goal of the training
phase is to learn weights for each features-cluster. At test time, each
extracted feature will be weighted by its cluster weight, and incorporated in
the search for the maximum sub-interval sum. If the sum at time $t_i$ 
is above a learned threshold $\theta_c$ and starts to decrease, then a detection
is triggered for action class $c$.}
\label{fig:pipeline}
\end{figure*}
%
%--------------------------------------------------------------------
%
\subsection{Our local descriptor}
\label{sec:descriptor}
The type of local descriptor has a direct impact on the recognition performance. 
We experimented with different descriptors, and as a byproduct contribution, 
we achieved best results with a 
proposed local descriptor that is a weighted concatenation of the angles 
descriptor~\cite{anglesDesc} and a slight modification of the Moving Pose 
descriptor~\cite{zanfir2013moving}\footnote{We modify the descriptor by rescaling the vector of 
concatenated joint positions to unit norm.}. 
The angles descriptor uses 35 angles 
between triplets of joints, $\Theta (t_0)$, along with angular velocity, $\delta \Theta (t_0)$.
On the other hand, the Moving Pose descriptor relies on joints positions, $P(t_0)$, relative to 
a reference joint, namely the hip 
center. And to capture kinematic information, it includes the first and second 
order derivatives; $\delta P(t_0)$ and $\delta ^2 P(t_0)$. 
The final form of our descriptor is $[[P, 
\alpha \delta P, \beta \delta^2P]$~$|$~$\psi [\Theta, \delta\Theta]]$, where 
$\alpha$ and $\beta$ are parameters defined in ~\cite{zanfir2013moving}, and 
$\psi$ is a weighting parameter to the relative importance of the two 
concatenated descriptors. In our experiments in section~\ref{sec:experiments}, 
we show that using our simple descriptor and with %simple
basic dictionary learning, 
we can achieve state-of-the-art results over different datasets.
%
% ===================================================================
%
\section{Efficient Linear Search (ELS)}
\label{sec:els}
In this section, we first explain our approach on offline action detection. Then,
we show how it can be easily extended to work for online detection as well. 
Figure~\ref{fig:pipeline} gives an overview of the approach.

Although many successful skeleton-based action recognition systems exist, most
of them haven't been extended to action detection. Sliding window approaches 
can be used for this task, evaluating the classifier function over all 
candidate sub-intervals.
So, for a video sequence S, it will identify the 
sub-interval, $s_{action} = [t_{st}, t_{end}]$, for which the classifier 
function produces the maximum score.
\begin{equation}\label{eq:maxSubwindow}
  s_{action} = argmax_{s \subseteq S} f(s)
\end{equation}
This identifies only one occurrence of the target action in the sequence. If 
multiple occurrences are to be found, then we can simply remove sub-intervals 
corresponding to previously identified actions, and repeat the search for the 
next $s_{max}$.
However, for a sequence of $N$ frames, we have $O(N^2)$ candidate sub-intervals, 
which incurs significant computational complexity. 
Thanks to Lampert \emph{et al.}~\cite{ess}, an efficient branch-and-bound method 
was proposed to search for the optimal bounding box of an object in an image. A 
limitation to branch-and-bound approaches is that they typically work offline, 
requiring the whole search space beforehand, as in~\cite{ess,yuan2011discriminative}.
In~\cite{yuan2011discriminative}, Yuan \emph{et al.} proposed a direct extension
of~\cite{ess} on action detection from RGB videos. They used a 
bag-of-features model based on spatio-temporal features, and proposed an offline
action detection system.
Another limitation to~\cite{ess,yuan2011discriminative} is that, optimizing equation~\ref{eq:maxSubwindow} 
may not always produce the desired 
behavior for a detection procedure since the optimal interval may contain multiple 
consecutive instances of an action. This problem is less likely to happen in the case 
of 2D images, for which the original branch-and-bound approach was first introduced~\cite{ess}.

In the following, we present a specialization of the branch-and-bound approach to the case of 
%action detection based on skeleton data. 
% skeleton-based action detection.
% \hl{skeleton-based action detection.}
skeleton-based action detection.
This turns out to be easily cast as one of the well-known 
dynamic programming problems that can be solved in linear time. Next, we show that a greedy 
approximation can effectively address the offline limitation of the branch-and-bound approach, 
as well as the problem of combining consecutive actions.
We assume two conditions: (1) a bag-of-gesturelets representation of action 
sequences, and (2) a linear binary classifier with good recognition accuracy 
trained for a specific target action.
While the linear classifier constraint is not necessary, as we later show, %TODO: don't forget to show it later..
we will use this assumption for simplicity of explanation.

%--------------------------------------------------------------------
\begin{figure*}[t]
        \centering
        %\captionsetup{justification=centering}
        \begin{subfigure}[b]{0.43\linewidth}
		\captionsetup{justification=centering}
                \includegraphics[width=0.9\textwidth]{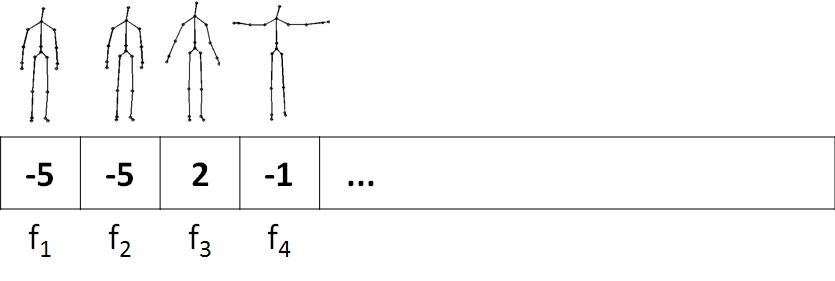}
                \caption{\small max subarray ending at f$_4$= [f$_{3}$, f$_{4}$]. Sum $= 1 < 
\theta_c$}
        \end{subfigure}%
        ~
        %add desired spacing between images, e. g. ~, \quad, \qquad, \hfill (or a blank line to force the subfigure onto a new line)
	\begin{subfigure}[b]{0.43\linewidth}
	      \captionsetup{justification=centering}
	      \includegraphics[width=0.9\textwidth]{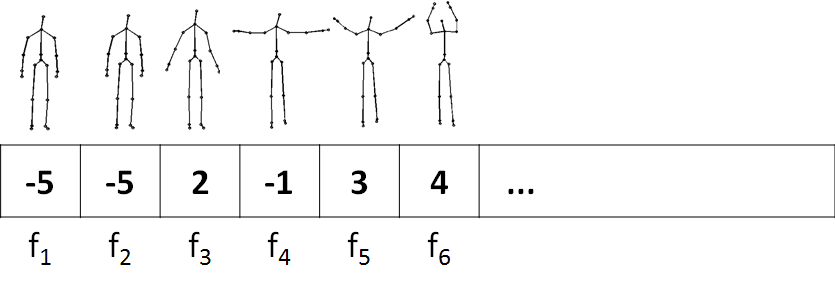}
	      \caption{\small sum $= 8 > \theta_c$. Action is being performed}
        \end{subfigure}

	%\captionsetup{justification=centering}
        \begin{subfigure}[b]{0.43\linewidth}
	      \captionsetup{justification=centering}
	      \includegraphics[width=0.9\textwidth]{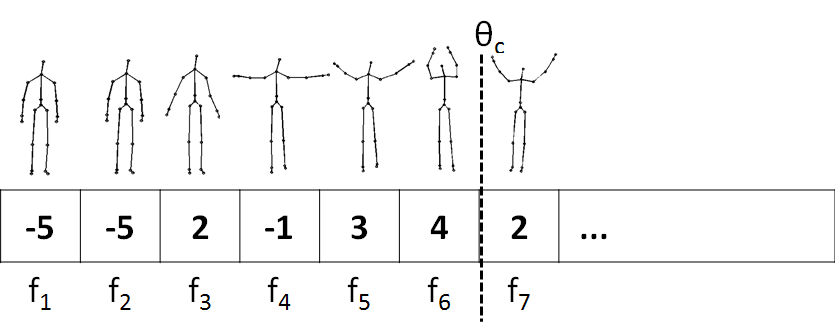}
	      \caption{\small max sum still increases. Action is still ongoing.}
        \end{subfigure}%
        ~
        %add desired spacing between images, e. g. ~, \quad, \qquad, \hfill (or a blank line to force the subfigure onto a new line)
	\begin{subfigure}[b]{0.43\linewidth}
	      \captionsetup{justification=centering}
	      \includegraphics[width=0.9\textwidth]{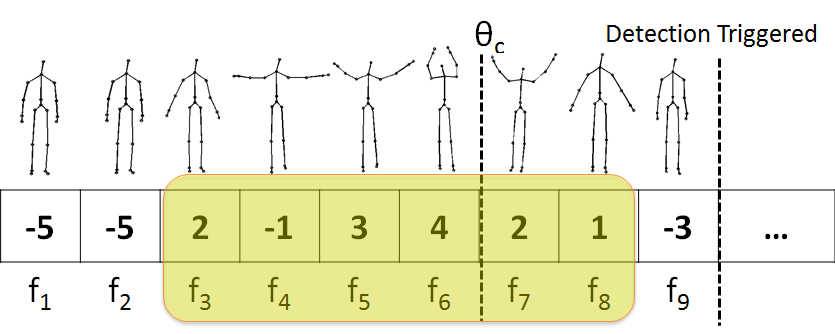}
	      \caption{\small max sum ($>\theta_c$) decreases. Trigger detection 
[f$_{3}$, f$_{8}$]}
        \end{subfigure}

        \caption{\small Example of online detection. We assume the action threshold
$\theta_c = 7$. Detected interval is [f$_{3}$, f$_{8}$].
% (figure is better seen in zoom)
}
        \label{fig:detection}
%\end{figure}
\end{figure*}
%
%--------------------------------------------------------------------
% 
\subsection{Offline action detection}
For any linear classifier, the corresponding scoring and decision functions take the form of:
\begin{equation} \label{eqn:linearClassifier}
 f(S) = w^T x + w_0, \quad 
    y(S) = \begin{cases}
      1 & \text{if } f(S) > 0\\
      0       & \text{otherwise}
      \end{cases}	
\end{equation}
where $S$ is a test sequence or sub-sequence,  $x$ is the 
feature vector of $S$, $w$ is the weight vector learned by the classifier, and 
$w_0$ is a constant bias. With the linearity of the dot product, $w^T x$, and the 
fact that $x$ is a histogram that counts the occurrence of each cluster index, 
the scoring function~\ref{eqn:linearClassifier} can be rewritten as:
\begin{equation}\label{eq:svmForm2}
  f(S) = w_0 + \sum_{j=1}^n w_{c_j}
\end{equation}
where, $c_j$ is the cluster index to which gesturelet $x_j$ belongs, and $n$ 
is the total number of gesturelets extracted from sequence $S$.~\footnote{For simplicity of presentation, 
this formulation does not consider soft binning in histogram construction. 
Including soft binning is straight forward though.} We can then 
evaluate the classifier function over any sub-sequence $s \subseteq S$ by summing the 
weights, $w_{c_j}$, of gesturelets $x_j$ that only belong to the sub-sequence 
$s$.
Since we want to identify the sub-sequence $s \subseteq S$ that maximizes 
equation~(\ref{eq:svmForm2}), we can safely drop the bias term, $w_0$.

The offline detection procedure will then be as follows: construct an empty 1D 
score array, with length equal to the number of frames.
This array will represent the per-point contribution of all extracted features 
from test sequence $S$. For each feature $x_j$, identify its cluster index 
$c_j$, then add up its weight, $w_{c_j}$, to the frame index in the score array. 
Finding the start and end frames of the action, $s_{action} = [t_{st}, 
t_{end}]$, that satisfy equation~\ref{eq:maxSubwindow} can now be mapped to 
finding the maximum subarray sum in the score array. Thanks to Kadane's 
algorithm~\cite{bentley1984programming}, this can be done in linear time in the number of frames, using 
dynamic programming. If $f(s_{max}) \geq \theta_c$, where $\theta_c$ is a learned 
threshold for action class $c$, then we specify that action $c$ has occurred, 
and we return the start and end frames of the sub-sequence $s_{max}$. 
To learn the action threshold for each action class, we compute the score of all 
training sequences, and search for the score threshold, $\theta_c$, that minimizes the binary 
classification error on the training data for class $c$.

%TODO: needs details and clarification...
To detect different action classes, we will construct a one-versus-all 
binary classification model for each action class. What differs from one action class 
to another is the classification model weights assigned to each entry in 
the codebook, along with the learned threshold $\theta_c$. 
The detection procedure for each of the classes can then be run concurrently on the action 
sequence with different per-point feature weights for each action class. 
% If one action class is detected at time $t_{i}$,
% then the detection procedure for all other classes is reset, and the search for a new detection
% starts from time $t_{i+1}$.

It is noted that, while the linearity of the classifier is a sufficient condition, 
it is not necessary. Any type of classification method that can output a classification score at the frame 
level can be used with our approach, since we later learn the score threshold, $\theta_c$, explicitly. This
gives flexibility to our approach with a wider range of classifiers.
%
%--------------------------------------------------------------------
% 
\subsection{Online action detection}
\label{sec:onlinedetection}
In this section we extend our approach to handle online action detection.
To find the maximum subarray sum, Kadane's algorithm scans through the
array. 
At each position (frame), it identifies the subinterval 
ending at this position with the maximum sum.
Since we only trigger the action if the 
maximum sum is larger
than a learned threshold $\theta_c$, then at any time $t_i$, if the maximum sum 
ending at frame $i$ exceeds $\theta_c$, then we
know that action $c$ is being performed, and our task, then, is to find at 
which frame the action will end. We expect that after the action ends, we will 
encounter a contiguous sequence of frames with negative scores. So, for the next 
frames, as long as the maximum sum grows, the action is still on-going. Once the 
subarray sum starts to decrease, we can assume that the action has ended. 
There will be a compromise between latency and confidence. If we specify the
end of action at the first negative point score, this will be very sensitive
to noise but it will achieve low latency. If we wait till we encounter
two consecutive points with negative scores, then this will achieve higher
confidence at the expense of increased latency, and so on. In our
experiments, we found that triggering the action at the first frame with negative score, 
after exceeding the threshold $\theta_c$,
causes a slight decrease of accuracy, while achieving very low latency.
Figure \ref{fig:detection} shows an example of how online detection works.

As we hinted before, a problem to the offline detection procedure is that it may combine 
consecutive repetitions of the same action in one detection. 
If an action is repeated twice, with a short pause between them, we expect that each repetition will
yield a high positive response, while the pause between them will have a negative response, but not as high.
So, optimizing equation~\ref{eq:maxSubwindow} would combine the two 
repetitions together. To separate the two repetitions, the separating negative score must be high. This
problem is solved in online detection with the greedy approach to terminating the action, which only requires consecutive
frames with negative scores, without further restrictions on their weight.
Section~\ref{sec:offlinevsonline} compares offline and online detection to highlight this offline detection limitation.

Detecting multiple classes in the online case is similar to the offline case, except that
if one action class is detected at time $t_{i}$, then the detection procedure for all other 
classes is reset, and the search for a new detection starts from time $t_{i+1}$.
% 
% ===================================================================
% 
\section{Experimental results}
\label{sec:experiments}
In this section, we first describe the two datasets used in our experiments.
We then expand on our implementation details.
Then, we compare the performance of our approach to the state of the art on both datasets.
% \footnote{Before the camera ready version, we found~\cite{wu2014leveraging} who report better results than ours on MSRC-12.} 
We start with the performance on the more classical action recognition task, then, we move to online action detection.
Next, we compare offline and online detections.
After that, we show how the performance of our approach is affected by changing its main parameters. 
Finally, we demonstrate its real-time performance.
%
%--------------------------------------------------------------------
% 
\subsection{Datasets}
\textbf{MSRC-12:}\quad %\hspace{0.2cm}
The Microsoft Research Cambridge-12 dataset~\cite{chi} is a large dataset designed for action detection. 
It contains more than $700,000$ frames
in 594 unsegmented sequences, encompassing $6,244$ gesture instances, recorded for 30 subjects 
performing 12 different gestures.
The samples consist of the 3D positions of 20 joints of the body skeleton captured using the Microsoft 
Kinect sensor at 30 fps.
The MSRC-12 dataset is annotated using the notion of an Action Point, which is a pose 
within the gesture that clearly identifies its completion.\\

\textbf{MSR-Action3D:}\quad %\hspace{0.2cm}
MSR-Action3D dataset~\cite{li2010action} is a standard dataset for action 
recognition.
It consists of $557$ pre-segmented sequences, with more than $20,000$ frames. 
There are 
10 subjects performing 20 different action gestures. Similar to MSRC-12, the 3D 
positions of 20 joints
are captured using the Microsoft Kinect sensor.
%
%--------------------------------------------------------------------
% 
\subsection{Implementation details}
\label{sec:implementationdetails}
For the linear classifier through our experiments, we use an SVM~\cite{CC01a} with a linear kernel.
The local descriptor has three parameters, $\alpha, \beta$, and $\psi$. While $\alpha, \beta$ are inherited 
from the Moving Pose descriptor~\cite{zanfir2013moving}, we introduced $\psi$ to weight the relative importance of the 
two concatenated descriptors.
Coarse-grain values for the parameters were learned from the training 
set, using different combinations of the 3 parameters in a brute-force manner. Trial-and-error was then used to fine-tune
the parameters' values.
For MSR-Action3D, we split the training set, persons $\{1,3,5,7,9\}$, into training and validation sets, using $40\%$ of the
training data (persons 7 and 9) as a validation set. Learned parameters ($\alpha=1, \beta=1, \psi=1.7$) were then 
fixed for all test experiments on MSR-Action3D. 
For MSRC-12, the parameters were learned over one modality ($20\%$ of the 
dataset), namely the video modality. Learned values
($\alpha=0.375, \beta=0.3, \psi=0.2$) were then used in test experiments over 
all modalities. For soft-binning, we set the number of neighbors, $m$, to 3. The vote to the $i^{th}$ nearest cluster
is weighted by $1/i$. A control experiment for choosing $m$ is shown in section~\ref{sec:sensitivity}.
 
The unsegmented sequences for the MSRC-12 dataset contain pauses between consecutive action instances, in which
the actor often stands still in a neutral pose. 
Such a neutral pose also occurs at the beginning and ending of most action instances.
Therefore, the neutral pose does not discriminate between different action classes, and hence, the 
classifier may be tempted to give it a neutral weight, possibly positive in sign. 
This would cause a problem for our detection procedure, which relies on having negative scores right before and 
after an action instance.
To overcome this problem, we add hard negatives to the negative training samples used to train our binary classifiers. 
Each of these hard negatives consists of one positive instance followed by a pause between two consecutive instances. 
In this way, the classifier is forced to give strong negative scores to the neutral poses in order to 
discriminate between those hard negatives and the positive samples, which solves the issue for the detection procedure.
A similar problem occurs in the detection on unsegmented sequences from the MSR-Action3D dataset. 
Actions start and end with neutral poses, which could again cause issues in localizing 
the beginning and ending of action instances. 
In this case, we include concatenations of two action instances as hard negatives in the binary classifier training.

A weighted moving-average on the frames scores was applied, where the anchor frame had weight equal to number of its neighbors.
For MSRC-12, we used a window of 5 frames, anchored at the middle frame (just as in the local descriptor). 
For MSR-Action, we used a window of 3 frames, since MSR-Action3D sequences are much smaller than those of MSRC-12.

\begin{table}
% \small
\begin{center}
    \begin{tabular}{|c|c|}
    \hline
      Method & Accuracy \\ \hline
      Eigenjoints~\cite{yang2012eigenjoints} & $82.3\%$ \\ \hline
      Random Occupy Pattern~\cite{wang2012robust} & $86.2\%$ \\ \hline
      Actionlets Ensemble~\cite{wang2012mining} & $88.2\%$ \\ \hline
      Covariance Descriptor (Cov3DJ)~\cite{hussein2013human} & $90.5\%$ \\ \hline
      Angles Covariance Descriptor~\cite{sharaf2014real} & $91.1\%$ \\ \hline
      \parbox[t]{5cm}{\centering Histogram of Oriented \\Displacements
      (HOD)~\cite{gowayyed2013histogram}} & $91.26\%$ \\ \hline
      Fusing Spatiotemporal Features~\cite{zhu2013fusing} & $94.3\%$ \\ \hline
      \parbox[t]{5.5cm}{\centering {\small Group Sparsity and Geometry Constrained 
      %TODO tooo looong and spanning 3 lines!!!
      Dictionary Learning (DL-GSGC)}~\cite{luo2013group}} & $\mathbf{97.27\%}$ \\ \hline \hline
      \textbf{Our Approach} & {\small $\mathbf{96.05 \pm 0.39 \%}$} \\ \hline
    \end{tabular}
\end{center}
\caption{\small Comparative recognition results on MSR-Action3D.}
\label{table:action3DRecognition}
\end{table}
%--------------------------------------------------------------------

\begin{table}
\begin{center}
    \begin{tabular}{|c|c|}
    \hline
      Method & Accuracy \\ \hline
      Cov3DJ Descriptor~\cite{hussein2013human} & $94.48\%$ \\ \hline
      Our Approach & $96.83\%$ \\ \hline
    \end{tabular}
\end{center}
\caption{\small Recognition results on MSRC-12 dataset.}
\label{table:msrc12Recognition}
\end{table}
%--------------------------------------------------------------------

\begin{table*}[ht!]
\begin{center}
    \begin{tabular}{|c|c|c|c|}

\hline

                 & Fothergill \emph{et al.}\cite{chi} & Sharaf \emph{et 
al.}\cite{sharaf2014real} & 
                 \multicolumn{1}{p{2.5cm}|}{\centering ELS} \\ \hline
    Video - Text & $0.679 \pm 0.035$ & $0.713 \pm 0.105$  & $0.790 \pm 0.133$ \\ \hline
    Image - Text & $0.563 \pm 0.045$ & $0.656 \pm 0.122$  & $0.711 \pm 0.228$ \\ \hline
    Text         & $0.479 \pm 0.104$ & $0.521 \pm 0.072$  & $0.622 \pm 0.246$ \\ \hline
    Video        & $0.627 \pm 0.052$ & $0.635 \pm 0.075$  & $0.726 \pm 0.225$ \\ \hline
    Image        & $0.549 \pm 0.102$ & $0.596 \pm 0.103$  & $0.670 \pm 0.254$ \\ \hline \hline
    Overall        & 0.579  & 0.624  & 0.704 \\ \hline

    \end{tabular}

\end{center}

\vspace{-5pt}

	\caption{\small Detection experiment for MSRC-12 dataset at $\bigtriangleup = 
333ms$ latency. Mean F-score and its standard deviation is reported for each instruction modality.}
	\label{table:MSRC12results}

\end{table*}

%
%--------------------------------------------------------------------
% 
\subsection{Action recognition}
\label{sec:exp_recognition}
First, to demonstrate the discriminative power of the BoG representation~\ref{sec:representation}, 
we report 
recognition results on both MSR-Action3D and MSRC-12 datasets. For 
MSR-Action3D, we follow 
the same experimental setup as~\cite{li2010action}; dividing the dataset into 3 
action sets and 
training with sequences performed by subjects $\{1,3,5,7,9\}$. Average accuracy 
over the 3 action sets is then reported.
We report the average classification rate and its standard deviation over 5 runs 
with 5 different codebooks.
Table~\ref{table:action3DRecognition} compares our results to state-of-the-art 
approaches.
As results show, the only approach that outperforms ours
is DL-GSGC~\cite{luo2013group}, where a dictionary learning algorithm is 
proposed for sparse coding. DL-GSGC adds group sparsity 
and geometric constraints to reconstruct feature points with minimal error. On 
the other hand, we achieve very competitive
results using basic dictionary learning with simple k-means clustering.

We also report classification results on MSRC-12 dataset.
Since this dataset is originally unsegmented
and labeled with action points only, we use the annotation and 
experimental setup of~\cite{hussein2013human} 
and compare our results to theirs in table~\ref{table:msrc12Recognition}. It is 
noted that, in~\cite{hussein2013human}, 
the covariance descriptor is used as a global descriptor constructed for the 
entire action sequence, 
while we use a local features approach to represent the action sequence. This 
signifies the power of using local pose
information in addition to joints kinematics information, instead of encoding 
global kinematic information 
only as in~\cite{hussein2013human}.
% 
%--------------------------------------------------------------------
% 
\begin{table*}[ht!]
  \begin{center}
    \begin{tabular}{|c|c|c|c|}

  \hline
		  & \parbox[t]{2.1cm}{\centering Sharaf \emph{et al.}\\ at $0.2$ overlap}\cite{sharaf2014real} & 
		  \parbox[t]{2.1cm}{\centering ELS\\ at $0.2$ overlap} & 
		  \parbox[t]{2.1cm}{\centering ELS\\ at $0.5$ overlap}\\ \hline
    Video - Text & $0.684 \pm 0.074$ & $0.921 \pm 0.126$ & $0.866 \pm 0.146$   \\ \hline
    Image - Text & $0.687 \pm 0.099$ & $0.894 \pm 0.085$ & $0.806 \pm 0.099$   \\ \hline
    Text         & $0.558 \pm 0.092$ & $0.788 \pm 0.139$ & $0.710 \pm 0.158$   \\ \hline
    Video        & $0.669 \pm 0.082$ & $0.895 \pm 0.068$ & $0.821 \pm 0.093$   \\ \hline
    Image        & $0.598 \pm 0.082$ & $0.858 \pm 0.086$ & $0.734 \pm 0.130$   \\ \hline \hline
    Overall        & $0.639$  & $0.871$ & $0.787$ \\ \hline

  \end{tabular}
\end{center}
\vspace{-5pt}
	\caption{\small Overlap detection experiment for MSRC-12 dataset at 0.2 and 0.5 overlap 
ratios. Mean F-score and its standard deviation is reported for each instruction modality.}
	\label{table:MSRC12Overlap}
\end{table*}

\subsection{Online action detection}

\subsubsection{Action detection on MSRC-12 dataset}

As mentioned before, MSRC-12 is annotated with action points.
So, most approaches experimenting on MSRC-12 (e.g.~\cite{chi, sharaf2014real}) 
convert the problem of 
real time action detection into classifying each frame as an action point or 
not, in real time. 
A positive detection is regarded when the gesture occurrence is detected within 
a short time window from the ground truth action point. 
Thanks to \cite{hussein2013human}, they provided manual annotation for the start 
and end 
frames of each gesture instance, which is needed to train classifiers for accurate action segmentation.
To be able to compare our results to \cite{chi, sharaf2014real}, we regard the 
gesture occurrence to begin at the start frame
as in the annotation of~\cite{hussein2013human}, and ends at the action point 
annotation. 
Our method, then, should trigger the occurrence of a gesture at its ground 
truth 
action point as in \cite{chi, sharaf2014real}.
Within each modality, we measure the precision and recall of each action class 
across all 10 folds. We then report the average F-score.

We use two different precision-recall experimental protocols.
First, a positive detection is counted if the detection is triggered within 10 
frames from the action point, this is a latency of 0.333 seconds as in 
\cite{chi, sharaf2014real}.
Table~\ref{table:MSRC12results} compares results of ELS against state-of-the-art 
results on MSRC-12.
\footnote{Before the camera ready version, we found~\cite{wu2014leveraging} which reports better results on MSRC-12.} 
%  \footnote{Before the CR version, we found~\cite{wu2014leveraging} who report better resuts.} 
It is noted that we don't compare with~\cite{sss}, as they use a different 
protocol for F-score calculation.

Second, we use the standard precision-recall experimental protocol used in 
object detection in images, which is an overlap threshold of 0.2, as 
in~\cite{duchenne2009automatic, zanfir2013moving}. This is to 
demonstrate the power of our approach for real-time segmentation of the temporal sequence, 
identifying 
the start and end of a gesture online and in real-time. 
Since we are the first to report overlap results on MSRC-12, we 
obtained the state-of-the-art code of Sharaf \emph{et al.}\cite{sharaf2014real} 
and reported 
its overlap results. We compare results in table~\ref{table:MSRC12Overlap}.
The results show that our approach 
significantly outperforms the state-of-the-art results in the overlap 
experiment. Our results for a $0.5$ overlap ratio are still even better than
Sharaf \emph{et al.} with a $0.2$ overlap ratio.
This emphasizes the power of our approach for real-time action segmentation.
It is noted that although \cite{sharaf2014real} uses a multi-scale approach, but 
it operates on the level of
the whole action sequence. On the other hand, the granularity of our
approach is a single frame with a small temporal window of 2 frames on each 
side.
Such finer granularity gives higher flexibility when identifying
the start and end of actions of different lengths and/or speed.
%
%--------------------------------------------------------------------
% 
\subsubsection{Action detection on MSR-Action3D dataset}
To further test our approach on another dataset, we conduct the same 
experiment of~\cite{zanfir2013moving} for online action detection on 
MSR-Action3D.
Since MSR-Action3D is designed for action recognition, where action sequences are 
pre-segmented,
\cite{zanfir2013moving} concatenates all test sequences in a random order to create one long 
unsegmented test sequence for action detection.
To be able to compare our results to~\cite{zanfir2013moving}, we train with 
persons $\{1,2,3,4,5\}$ and test with the rest.
We also repeat this experiment
100 times with different random concatenation ordering as done in~\cite{zanfir2013moving},
and compare our results in table~\ref{table:action3DDetection}. As results show, we are on the bar 
with state-of-the-art results 
on this dataset. It is, however, noted that a limitation 
to~\cite{zanfir2013moving} is using k-NN search on all training frames. This works well for small 
datasets, like MSR-Action3D. 
However, for large datasets, k-NN will potentially incur significant space and time requirements. 
On the other hand, the complexity of our approach is primarily affected by the codebook size, which is
significantly smaller than the number of frames in the training data.

% \begin{table}[H]
\begin{table}[h!]
  \begin{center}
    \begin{tabular}{ccc}
  \hline
    Method  &  Detection mean AP \\ \hline
    The Moving Pose\cite{zanfir2013moving} & $0.890 \pm 0.002$ \\ %\hline
    ELS & $0.902 \pm 0.007$ \\ \hline
  \end{tabular}
\end{center}
\vspace{-5pt}
	\caption{\small Detection experiment for MSR-Action3D at a 0.2 overlap ratio.}
	\label{table:action3DDetection}

\end{table}

\begin{figure*}[ht!]
        \centering
        \begin{subfigure}[b]{0.45\linewidth}
		\captionsetup{justification=centering}
                \includegraphics[width=0.87\textwidth]{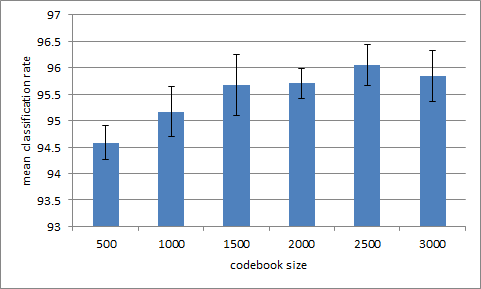}
                \caption{\small Recognition results vs codebook size for \\
MSR-Action3D}
		\label{subfig:controlCodebook}
        \end{subfigure}
        ~
	\begin{subfigure}[b]{0.45\linewidth}
		\captionsetup{justification=centering}
		\includegraphics[width=0.87\textwidth]{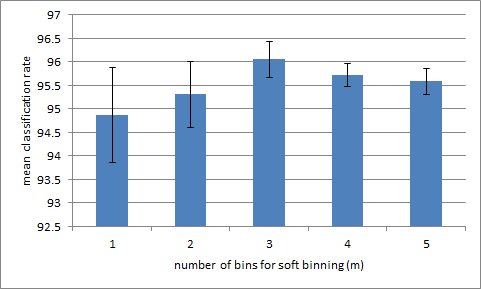}
		\caption{\small Recognition results vs number of neighbors, $m$,\\ for 
soft-binning}
	      \label{subfig:controlSoftBinning}
        \end{subfigure}
        \caption{\small Control Experiments}
\end{figure*}
%
%--------------------------------------------------------------------
% 
\subsection{Offline vs online detection}
\label{sec:offlinevsonline}
Table~\ref{table:offline} compares offline and online detection results on both MSRC-12 and MSR-Action3D datasets. 
It reports the detection F-score for a $0.2$ overlap ratio.
The results highlight the limitations of offline detection and how the online detection overcomes them,
as illustrated in~\ref{sec:onlinedetection}.
Since the unsegmented sequences in MSRC-12 are mainly repeated performances of the same action class,
the offline detection procedure first concatenated all repetitions in one long detection.
Although hard negatives~\ref{sec:implementationdetails} partially solved this problem, but still around $85\%$ of 
the gestures that failed to be detected, were missed due to concatenation to subsequent actions. 
For MSR-Action3D, around $64\%$ of actions that failed to be detected were missed due to partial or 
complete concatenation to subsequent
actions. For example, in the synthesized unsegmented test sequence, there were 14 occurrences where the same action
was repeated twice. While the online detection successfully separated and detected the two repeated actions
in 13 out of 14 occurrences, the offline detection
failed to separate the two repeated actions in all 14 occurrences.

\begin{table}[h!]
\begin{center}
    \begin{tabular}{|c|c|c|}
    \hline
      Dataset & Offline F-score & Online F-score\\ \hline
      MSRC-12 & $0.524$  & $0.871$ \\ \hline
      MSR-Action3D & $0.794$ & $0.930$  \\ \hline
    \end{tabular}
\end{center}
\caption{\small Offline and online detection F-score, for a $0.2$ overlap ratio.}
\label{table:offline}
\end{table}
% 
%--------------------------------------------------------------------
% 
\subsection{Sensitivity analysis}
\label{sec:sensitivity}
In this section we show the sensitivity of our framework's performance to changing different parameters. We 
conduct these experiments on the MSR-Action3D dataset since it is of reasonable size, which allows 
extensive experimentation. Without loss of generality, we conduct these experiments on the more basic action recognition task.
Since our results depend on the constructed codebook, whose construction involves randomness, we repeat each experiment 
5 times using 5 different codebooks and report the mean result and its standard deviation.

The first parameter is the choice of codebook size, $K$. 
The suitable codebook size is expected to be proportional to the target dataset size and the number of 
gesture classes in it. 
Figure~\ref{subfig:controlCodebook} shows recognition results on MSR-Action3D 
with different codebook sizes.
The results show that our framework is not sensitive to the 
choice of codebook size. It works well even in the case of small codebooks. 
Also, the small standard 
deviation means that the clustering randomness in codebook construction is of 
minor effect.
This indicates both the robustness of our approach, and the discriminative power 
of our local descriptor.

Next, we show the effect of using soft binning~\ref{sec:representation} with 
different choices of $m$, where
$m$ is the number of neighbors for which each feature casts a weighted vote. For 
this experiment, we fix
the codebook size to 2500, and plot recognition result vs.~different settings of 
$m$ 
in figure~\ref{subfig:controlSoftBinning}. Although recognition accuracy 
improves in the case of using
soft binning, the more important effect is evident in how the standard deviation 
significantly reduces
with an increased value of $m$. This is because, with soft binning, it is less 
likely to miss voting for 
the correct representative cluster, unlike the hard assignment to one cluster only. 

Last, we show the effect of using different local descriptors with our approach, in 
table~\ref{table:features}. We achieve best results using our proposed descriptor~\ref{sec:descriptor}.

%--------------------------------------------------------------------

% \begin{table}[H]
\begin{table}[h!]
{
  \begin{center}
    \begin{tabular}{ccc}
  \hline
    Used descriptor  &  Accuracy \\ \hline
    Angles Covariance~\cite{sharaf2014real} & $83.93\%$ \\
    Angles Descriptor~\cite{anglesDesc} & $86.25\%$ \\
    The Moving Pose~\cite{zanfir2013moving} & $94.01\%$ \\ %\hline
    Our proposed descriptor & $96.05\%$ \\ \hline
  \end{tabular}
\end{center}
\vspace{-5pt}
	\caption{\small Effect of using different features in our approach on 
MSR-Action3D recognition}
	\label{table:features}
} 
\end{table}
% 
%--------------------------------------------------------------------
% 
\subsection{Real-time operation}
The goal of our framework is not only to perform online action detection, but 
also to do this in real-time. There are 3 main factors affecting the running 
time of our approach: 1) Size of the codebook; 2) Size of the local 
descriptor. 3) Number of neighbors, $m$, to vote for, in soft binning.
For this experiment, we used a large codebook of size 3000. The dimensionality
of our descriptor is 250, and we used $m=3$.
The average running-time per frame of our MATLAB implementation
% \footnote{\hl{Matlab source code is available on the authors' webpages.}}
\footnote{Matlab source code is available on the authors' webpages.}
% \footnote{Source code and used codebooks are available on the author's webpage.} 
% \footnote{Code along with used codebooks will be released upon acceptance.}
was measured to be $10.7ms$. 
So, our framework can process approximately $93$ frames per second. The running-time 
was measured on a machine with 2.2 GHz Intel quad-core Core-i7 processor and 12 
GB RAM. 
% 
% ===================================================================
% 
\section{Conclusion}
\label{sec:conclusion}
We have proposed both a simple skeleton-based descriptor and a novel approach for action detection. 
The proposed approach maximizes a binary classifier score over all possible sub-sequences, typically in linear time.
It can be used in conjunction with a large class of classifiers and with any local descriptor. 
Our proposed approach works online and at real-time with low latency.
It detects a gesture by specifying its start and end frames in an unsegmented video
sequence, which makes it suitable for real-time video temporal segmentation.
While the proposed method relies on simple components, we showed that a specialization for skeleton-based
action detection can be established which, not only outperforms the state-of-the-art, but also overcomes 
the limitations of similar approaches.
% 
% ===================================================================

\section{Acknowledgement}
% \hl{The authors would like to thank SmartCI Research Center for supporting this research.}
The authors would like to thank SmartCI Research Center for supporting this research.

% ===================================================================
% 
\pagebreak
{\small
\bibliographystyle{ieee}
\bibliography{refs}
}

\end{document}